\documentclass[10pt,twocolumn,letterpaper]{article}

\usepackage{cvpr}
\usepackage{times}
\usepackage{epsfig}
\usepackage{graphicx}
\usepackage{amsmath}
\usepackage{amssymb}
\usepackage{booktabs}
\usepackage{threeparttablex}
\usepackage{xspace}
\usepackage{multirow}
\usepackage{bbding}
\usepackage{multicol}
\usepackage{appendix}
\usepackage{widetext}

\usepackage{listings}
\usepackage{color}
\usepackage{colortbl}
\definecolor{mygray}{gray}{.9}

\definecolor{dkgreen}{rgb}{0,0.6,0}
\definecolor{gray}{rgb}{0.5,0.5,0.5}
\definecolor{mauve}{rgb}{0.58,0,0.82}
\usepackage[breaklinks=true,bookmarks=false]{hyperref}

\cvprfinalcopy 

\def\cvprPaperID{****} 

\begin{document}
\hyphenpenalty=2000
\tolerance=3000

\title{TEA: Temporal Excitation and Aggregation for Action Recognition}


\author{
Yan Li\textsuperscript{1}
~~~~
Bin Ji\textsuperscript{2}
~~~~
Xintian Shi\textsuperscript{1}
~~~~
Jianguo Zhang\textsuperscript{3~\thanks{Corresponding authors.}}
~~~~
Bin Kang\textsuperscript{1~\footnotemark[1]}
~~~~
Limin Wang\textsuperscript{2~\footnotemark[1]}\\
\textsuperscript{1} Platform and Content Group (PCG), Tencent\\
\textsuperscript{2} State Key Laboratory for Novel Software Technology, Nanjing University, China\\
\textsuperscript{3} Department of Computer Science and Engineering, Southern University of Science and Technology, China\\
{\tt\small phoenixyli@tencent.com, binjinju@smail.nju.edu.cn, tinaxtshi@tencent.com}  \\
{\tt\small zhangjg@sustech.edu.cn, binkang@tencent.com, 07wanglimin@gmail.com}
}

\maketitle
\begin{abstract}
  Temporal modeling is key for action recognition in videos. It normally considers both \textbf{short}-range motions and \textbf{long}-range aggregations. In this paper, we propose a Temporal Excitation and Aggregation (TEA) block, including a motion excitation (ME) module and a multiple temporal aggregation (MTA) module, specifically designed to capture both short- and long-range temporal evolution. In particular, for short-range motion modeling, the ME module calculates the feature-level temporal differences from spatiotemporal features. It then utilizes the differences to excite the motion-sensitive channels of the features. The long-range temporal aggregations in previous works are typically achieved by stacking a large number of local temporal convolutions. Each convolution processes a local temporal window at a time. In contrast, the MTA module proposes to deform the local convolution to a group of sub-convolutions, forming a hierarchical residual architecture. Without introducing additional parameters, the features will be processed with a series of sub-convolutions, and each frame could complete multiple temporal aggregations with neighborhoods. The final equivalent receptive field of temporal dimension is accordingly enlarged, which is capable of modeling the long-range temporal relationship over distant frames. The two components of the TEA block are complementary in temporal modeling. Finally, our approach achieves impressive results at low FLOPs on several action recognition benchmarks, such as Kinetics, Something-Something, HMDB51, and UCF101, which confirms its effectiveness and efficiency.
\end{abstract}

\section{Introduction}

Action recognition is a fundamental problem in video-based tasks. It becomes increasingly demanding in video-based applications, such as intelligent surveillance, autonomous driving, personal recommendation, and entertainment \cite{poquet2018video}.
Though visual appearances (and its context) is important for action recognition, it is rather important to model the temporal structure. Temporal modeling normally presents (or is considered) at different scales: 1) \textit{short}-range motion between adjacent frames and 2) \textit{long}-range temporal aggregation at large scales. There are lines of works considering one or both of those aspects, especially in the current era of deep CNNs \cite{karpathy2014large,simonyan2014two,yue2015beyond,donahue2015long,tran2015learning,yao2015describing,sun2015human,bilen2016dynamic,bilen2017action,varol2017long,zhao2018trajectory,zhao2018recognize,ng2018temporal,wang2018appearance,qiu2019learning,li2019temporal,jiang2019stm}. Nevertheless, they still leave some gaps, and the problem is far from being solved, \ie, it remains unclear how to model the temporal structure with significant variations and complexities effectively and efficiently.

For short-range motion encoding, most of the existing methods \cite{simonyan2014two,wang2016temporal} extract hand-crafted optical flow \cite{zach2007duality} first, which is then fed into a 2D CNN-based two-stream framework for action recognition. Such a two-stream architecture processes RGB images and optical flow in each stream separately. The computation of optical flow is time-consuming and storage demanding. In particular, the learning of spatial and temporal features is isolated, and the fusion is performed only at the late layers. To address these issues, we propose a {\itshape motion excitation} (ME) module. Instead of adopting the pixel-level optical flow as an additional input modality and separating the training of temporal stream with the spatial stream, our module could integrate the motion modeling into the whole spatiotemporal feature learning approach. Concretely, the feature-level motion representations are firstly calculated between adjacent frames. These motion features are then utilized to produce modulation weights. Finally, the motion-sensitive information in the original features of frames can be excited with the weights. In this way, the networks are forced to discover and enhance the informative temporal features that capture differentiated information.

For long-range temporal aggregation, existing methods either 1) adopt 2D CNN backbones to extract frame-wise
features and then utilize a simple temporal max/average pooling to obtain the whole video representation \cite{wang2016temporal,girdhar2017actionvlad}. Such a simple summarization strategy, however, results in temporal information loss/confusion; or 2) adopt local 3D/(2+1)D convolutional operations to process local temporal window \cite{tran2015learning,carreira2017quo}. The long-range temporal relationship is indirectly modeled by repeatedly stacking local convolutions in deep networks. However, repeating a large number of local operations will lead to optimization difficulty \cite{he2016deep}, as the message needs to be propagated through the long path between distant frames. To tackle this problem, we introduce a {\itshape multiple temporal aggregation} (MTA) module. The MTA module also adopts (2+1)D convolutions, but a group of sub-convolutions replaces the 1D temporal convolution in MTA. The sub-convolutions formulate a hierarchical structure with residual connections between adjacent subsets. When the spatiotemporal features go through the module, the features realize multiple information exchanges with neighboring frames, and the equivalent temporal receptive field is thus increased multiple times to model long-range temporal dynamics.

The proposed ME module and MTA module are inserted into a standard ResNet block \cite{he2016deep,he2016identity} to build the Temporal Excitation and Aggregation (TEA) block, and the entire network is constructed by stacking multiple blocks. The obtained model is efficient: benefiting from the light-weight configurations, the FLOPs of the TEA network are controlled at a low level (only 1.06$\times$ as many as 2D ResNet). The proposed model is also effective: the two components of TEA are complementary and cooperate in endowing the network with both short- and long-range temporal modeling abilities. To summarize, the main contributions of our method are three-fold:

1. The motion excitation (ME) module to integrate the short-range motion modeling with the whole spatiotemporal feature learning approach.

2. The multiple temporal aggregation (MTA) module to efficiently enlarge the temporal receptive field for long-range temporal modeling.

3. The two proposed modules are both simple, light-weight, and can be easily integrated into standard ResNet block to cooperate for effective and efficient temporal modeling.
\section{Related Works}

With the tremendous success of deep learning methods on image-based recognition tasks \cite{krizhevsky2012imagenet,simonyan2014very,szegedy2015going,he2016deep,he2016identity}, some researchers started to explore the application of deep networks on video action recognition task \cite{karpathy2014large,simonyan2014two,tran2015learning,yue2015beyond,donahue2015long,yao2015describing}. Among them, Karpathy \etal \cite{karpathy2014large} proposed to apply a single 2D CNN model on each frame of videos independently and explored several strategies to fuse temporal information. However, the method does not consider the motion change between frames, and the final performance is inferior to the hand-crafted feature-based algorithms. Donahue \etal \cite{donahue2015long} used LSTM \cite{hochreiter1997long} to model the temporal relation by aggregating 2D CNN features. In this approach, the feature extraction of each frame is isolated, and only high-level 2D CNN features are considered for temporal relation learning.

The existing methods usually follow two approaches to improve temporal modeling ability. The first one was based on two-stream architecture proposed by Simonyan and Zisserman \cite{simonyan2014two}. The architecture contained a spatial 2D CNN that learns still feature from frames and a temporal 2D CNN that models motion information in the form of optical flow \cite{zach2007duality}. The training of the two streams is separated, and the final predictions for videos are averaged over two streams. Many following works had extended such a framework. \cite{feichtenhofer2016convolutional,feichtenhofer2017spatiotemporal} explored different mid-level combination strategies to fuse the features of two streams. TSN \cite{wang2016temporal} proposed the sparse sampling strategy to capture long-range video clips. All these methods require additional computation and storage costs to deal with optical flow. Moreover, the interactions between different frames and the two modalities are limited, which usually occur at late layers only. In contrast, our proposed method discards optical flow extraction and learns approximate feature-level motion representations by calculating temporal differences. The motion encoding can be integrated with the learning of spatiotemporal features and utilized to discover and enhance their motion-sensitive ingredients.

The most recent work STM \cite{jiang2019stm} also attempted to model feature-level motion features and inserts motion modeling into spatiotemporal feature learning. Our method differs from STM in that STM directly {\itshape adds} the spatiotemporal features and motion encoding together. In contrast, our method utilizes motion features to {\itshape recalibrate} the features to enhance the motion pattern.

\begin{figure*}[]
    \begin{center}
        \includegraphics[height=5.6cm]{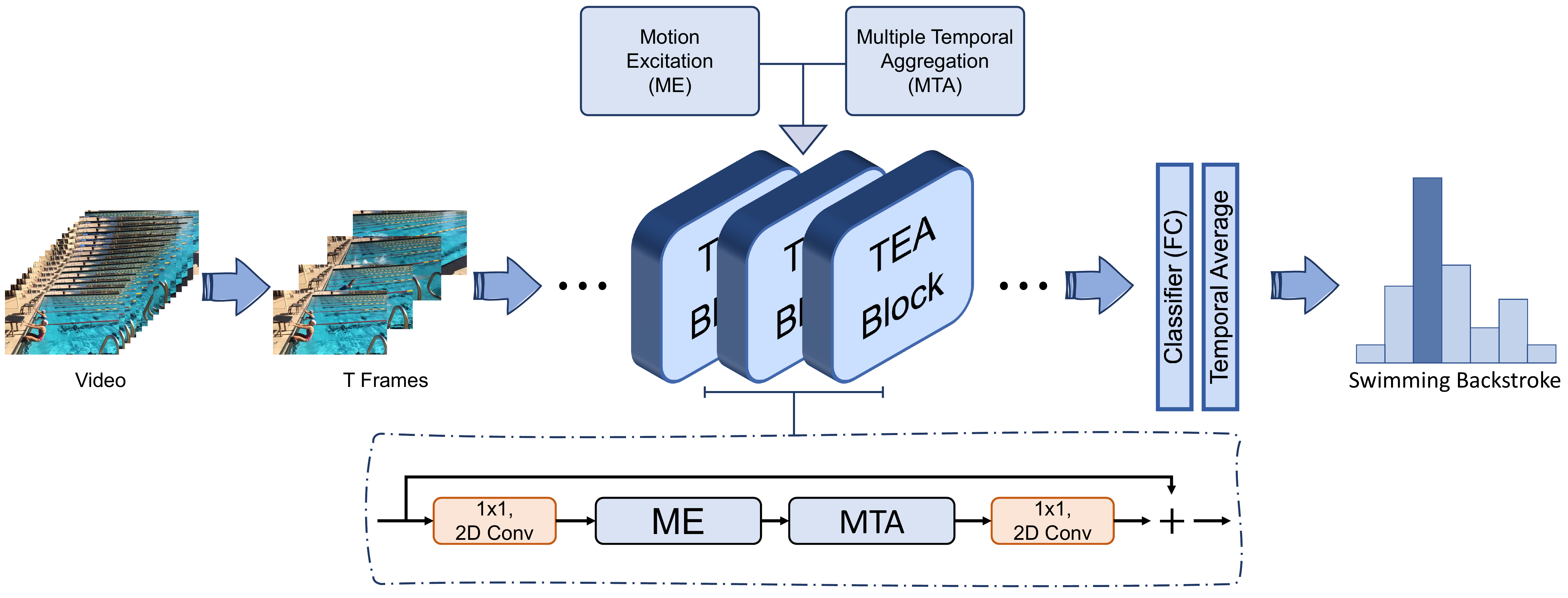}
    \end{center}
    \vspace{-1.5em}
        \caption{The framework of the proposed method for action recognition. The sparse sampling strategy \cite{wang2016temporal} is adopted to sample $T$ frames from videos. The 2D ResNet \cite{he2016deep} is utilized as the backbone, and the ME and MTA modules are inserted into each ResNet block to form the TEA block. The simple temporal pooling is applied to average action predictions for the entire video.}
    \label{fig:framework}
    \vspace{-1.5em}
    \end{figure*}

Another typical video action recognition approach is based on 3D CNNs and its (2+1)D CNN variants \cite{tran2015learning,sun2015human,carreira2017quo,tran2018closer,wang2018videos}. The first work in this line was C3D \cite{tran2015learning}, which performed 3D convolutions on adjacent frames to jointly model the spatial and temporal features in a unified approach. To utilize pre-trained 2D CNNs, Carreira and Zisserman \cite{carreira2017quo} proposed I3D to inflate the pre-trained 2D convolutions to 3D ones. To reduce the heavy computations of 3D CNNs, some works proposed to decompose the 3D convolution into a 2D spatial convolution and a 1D temporal convolution \cite{sun2015human,diba2017temporal,lin2019tsm,he2019stnet,qiu2019learning,tran2019video} or utilize a mixup of 2D CNN and 3D CNN \cite{tran2018closer,xie2018rethinking,zolfaghari2018eco}. In these methods, the long-range temporal connection can be theoretically established by stacking multiple local temporal convolutions. However, after a large number of local convolution operations, the useful features from distant frames have already been weakened and cannot be captured well. To address this issue, T3D \cite{diba2017temporal} proposed to adopt densely connected structure \cite{huang2017densely} and combined different temporal windows \cite{szegedy2015going}. Non-local module \cite{wang2018non} and stnet \cite{he2019stnet} applied self-attention mechanism to model long-range temporal relationship. Either additional parameters or time-consuming operations accompany these attempts. Different from these works, our proposed multiple temporal aggregation module is simple and efficient without introducing extra operators.

\section{Our Method}
The framework of the proposed method is illustrated in Figure \ref{fig:framework}. The input videos with variable lengths are sampled using the sparse temporal sampling strategy proposed by TSN \cite{wang2016temporal}. Firstly, the videos are evenly divided into $T$ segments. Then one frame is randomly selected from each segment to form the input sequence with $T$ frames. For spatiotemporal modeling, our model is based on 2D CNN ResNet \cite{he2016deep} and constructed by stacking multiple Temporal Excitation and Aggregation (TEA) blocks. The TEA block contains a motion excitation (ME) module to excite motion patterns and a multiple temporal aggregation (MTA) module to establish a long-range temporal relationship. Following previous methods \cite{wang2016temporal,lin2019tsm}, the simple temporal average pooling is utilized at the end of the model to average the predictions of all frames.

\begin{figure}[]
    \begin{center}
    \advance\leftskip -0.3cm
        \includegraphics[width=7.5cm]{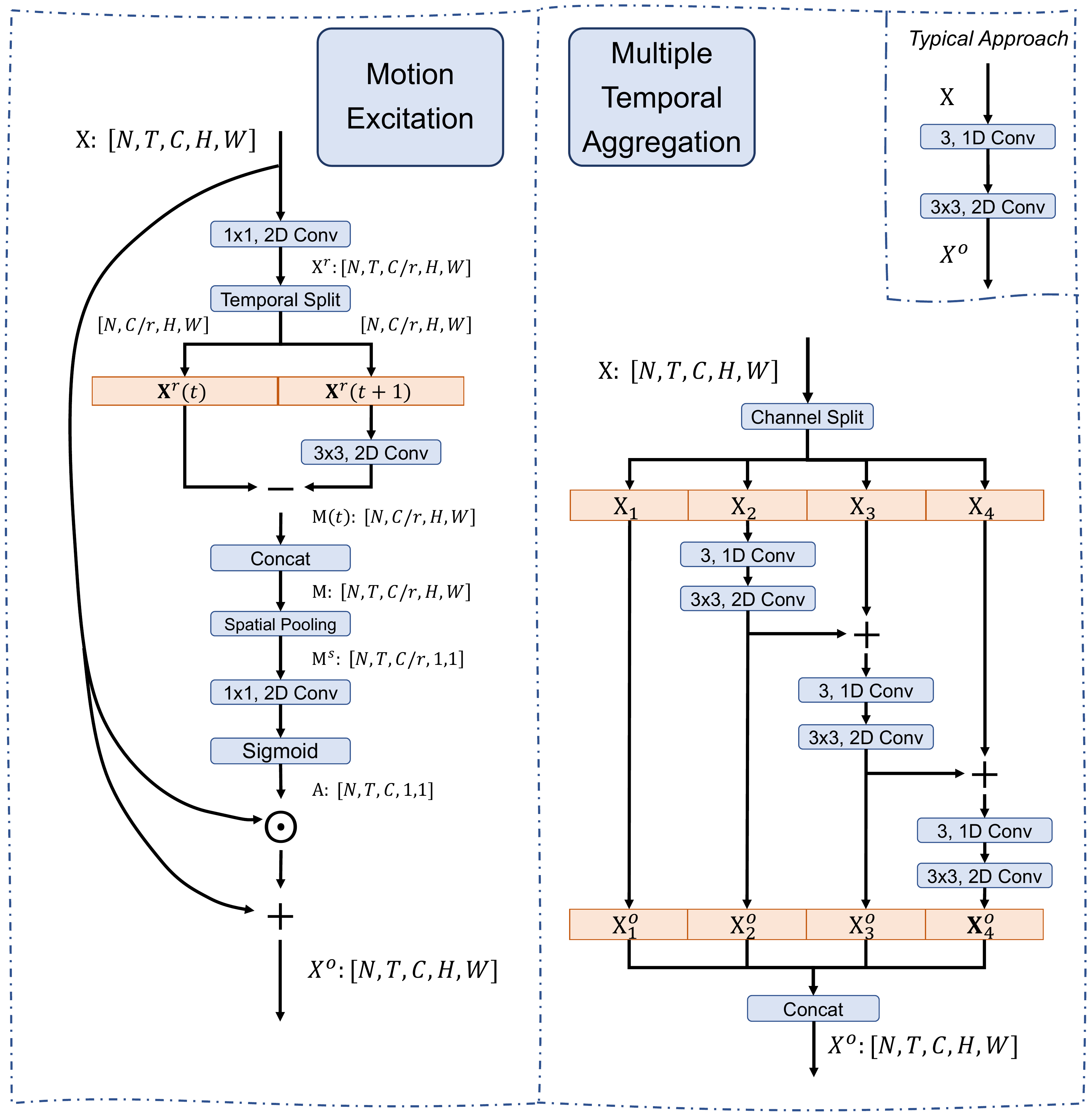}
    \end{center}
    \vspace{-1.5em}
        \caption{The implementations of the motion excitation (ME) module (left panel) and multiple temporal aggregation (MTA) module (right panel).}    \label{fig:meme}
            \vspace{-1.5em}
    \end{figure}
\subsection{Motion Excitation (ME) Module}
Motion measures the content displacements of the two successive frames and mainly reflects the actual actions. Many previous works utilize motion representations for action recognition \cite{wang2016temporal,carreira2017quo}. Still, most of them only consider {\itshape pixel-level} motion pattern in the form of optical flow \cite{zach2007duality} and separate the learning of motions from spatiotemporal features. Different from this, in the proposed {\itshape motion excitation} (ME) module, the motion modeling is extended from the raw {\itshape pixel-level} to a largely scoped {\itshape feature-level}, such that the motion modeling and spatiotemporal features learning are incorporated into a unified framework.

The architecture of the ME module is shown in the left panel of Figure \ref{fig:meme}. The shape of input spatiotemporal feature $\mathbf{X}$ is $\left[ N,T,C,H,W \right]$,  where $N$ is the batch size. $T$ and $C$ denote temporal dimension and feature channels, respectively. $H$ and $W$ correspond to spatial shape. The intuition of the proposed ME module is that, among all feature channels, different channels would capture distinct information. A portion of channels tends to model the static information related to background scenes; other channels mainly focus on dynamic motion patterns describing the temporal difference. For action recognition, it is beneficial to enable the model to discover and then enhance these motion-sensitive channels.

Given an input feature $\mathbf{X}$, a 1$\times$1 2D convolution layer is firstly adopted to reduce feature channels for efficiency.
\begin{equation}
	\mathbf{X}^r = \mathrm{conv}_\mathit{red} * \mathbf{X}, \quad \mathbf{X}^r \in \mathbb{R}^{N \times T \times C/r \times H \times W}
\end{equation}
where $\mathbf{X}^r$ denotes the channel-reduced feature. $*$ indicates the convolution operation. $r=16$ is the reduction ratio.

The feature-level motion representations at time step $t$ is approximately considered as the difference between the two adjacent frames, $\mathbf{X}^r(t)$ and $\mathbf{X}^r(t+1)$. Instead of directly subtracting the original features, we propose to perform the {\itshape channel-wise} transformation on features first and then utilize the transformed feature to calculate motions.
Formally,
\begin{equation}
	\mathbf{M}(t) = \mathrm{conv}_\mathit{trans} * \mathbf{X}^r(t+1) - \mathbf{X}^r(t), 1 \leq t \leq T-1,
\end{equation}
where $\mathbf{M}(t) \in \mathcal{R}^{N \times C/r \times H \times W}$ is the motion feature at time $t$. $\mathrm{conv}_\mathit{trans}$ is a 3$\times$3 2D channel-wise convolution layer performing transformation for each channel.

We denote the motion feature at the end of time steps as zero, \ie, $\mathbf{M}(T)=0$, and construct the final motion matrix $\mathbf{M}$ by concatenating all the motion features $\left[ \mathbf{M}(1), \dots, \mathbf{M}(T)\right]$. Then a global average pooling layer is utilized to summarize the spatial information since our goal is to excite the motion-sensitive channels where the detailed spatial layouts are of no great importance:
\begin{equation}
	\mathbf{M}^{s} = \mathrm{Pool}(\mathbf{M}), \quad \mathbf{M}^{s} \in \mathbb{R}^{N \times T \times C/r \times 1 \times 1}.
\end{equation}

Another 1$\times$1 2D convolution layer $\mathrm{conv}_\mathit{exp}$ is utilized to expand the channel dimension of motion features to the original channel dimension $C$, and the motion-attentive weights $\mathbf{A}$ can be obtained by using the sigmoid function.
\begin{equation}
	\mathbf{A} = 2\delta( \mathrm{conv}_\mathit{exp} * \mathbf{M}^s)-1, \quad \mathbf{A} \in \mathbb{R}^{N \times T \times C \times 1 \times 1},
\end{equation}
where $\delta$ indicates the sigmoid function.

Finally, the goal of the module is to excite the motion-sensitive channels; thus, a simple way is to conduct channel-wise multiplication between the input feature $\mathbf{X}$ and attentive weight $\mathbf{A}$. However, such an approach will suppress the static background scene information, which is also beneficial for action recognition. To address this issue, in the proposed motion-based excitation module, we propose to adopt a {\itshape residual} connection to enhance motion information meanwhile preserve scene information.
\begin{equation}
	\mathbf{X}^{o} = \mathbf{X} + \mathbf{X} \odot \mathbf{A}, \quad \mathbf{X}^{o} \in \mathbb{R}^{N \times T \times C \times H \times W},
\end{equation}
where $\mathbf{X}^{o}$ is the output of the proposed module, in which the motion pattern has been excited and enhanced. $\odot$ indicates the channel-wise multiplication.

\subsubsection{Discussion with SENet}
The excitation scheme is firstly proposed by SENet \cite{hu2018squeeze,hu2019squeeze} for image recognition tasks. We want to highlight our differences with SENet. 1) SENet is designed for image-based tasks. When SENet is applied to spatiotemporal features, it processes each frame of videos independently without considering temporal information. 2) SENet is a kind of self-gating mechanism \cite{wang2017residual}, and the obtained modulation weights are utilized to enhance the informative channels of feature $\mathbf{X}$. While our module aims to enhance the motion-sensitive ingredients of the feature. 3) The useless channels will be completely suppressed in SENet, but the static background information can be preserved in our module by introducing a residual connection.

\subsection{Multiple Temporal Aggregation (MTA) Module}

Previous action recognition methods \cite{tran2015learning,sun2015human} typically adopt the local temporal convolution to process neighboring frames at a time, and the long-range temporal structure can be modeled only in deep networks with a large number of stacked local operations. It is an ineffective approach since the optimization message delivered from distant frames has been dramatically weakened and cannot be well handled. To address this issue, we propose the {\itshape multiple temporal aggregation} (MTA) module for effective long-range temporal modeling. The MTA module is inspired by Res2Net \cite{gao2019res2net}, in which the spatiotemporal features and corresponding local convolution layers are split into a group of subsets. This approach is efficient since it does not introduce additional parameters and time-consuming operations. In the module, the subsets are formulated as a hierarchical residual architecture such that a serial of sub-convolutions are successively applied to the features and could accordingly enlarge the equivalent receptive field of the temporal dimension.

As shown in the upper-right corner of Figure \ref{fig:meme}, given an input feature $\mathbf{X}$, a typical approach is to process it with a single local temporal convolution and another spatial convolution. Different from this, we split the feature into four fragments along the channel dimension, and the shape of each fragment thus becomes $\left[ N,T,C/4,H,W\right]$. The local convolutions are also divided into multiple sub ones. The last three fragments are sequentially processed with one {\itshape channel-wise} temporal sub-convolution layer and another spatial sub-convolution layer. Each of them only has 1/4 parameters as original ones. Moreover, the residual connection is added between the two adjacent fragments, which transforms the module from a parallel architecture to a hierarchical cascade one. Formally\footnote{The necessary reshape and permutation operations are ignored for simplicity. In fact, to conduct 1D temporal convolution on input feature $\mathbf{X}$, it requires to be reshaped from $\left[ N,T,C,H,W\right]$ to $\left[ NHW, C, T\right]$.},
\begin{equation}
	\begin{array}{lr}
	    \mathbf{X}^o_i = \mathbf{X}_i, & i=1, \\
	    \mathbf{X}^o_i = \mathrm{conv}_{\mathit{spa}}* ( \mathrm{conv}_{\mathit{temp}} * \mathbf{X}_i), & i=2, \\
	    \mathbf{X}^o_i = \mathrm{conv}_{\mathit{spa}}* ( \mathrm{conv}_{\mathit{temp}} * ( \mathbf{X}_i + \mathbf{X}^o_{i-1} ) ), & i=3,4, \\
	\end{array}
\end{equation}
where $\mathbf{X}^o_i \in \mathbb{R}^{N \times T \times C/4 \times H \times W}$ is the output of $i$-th fragment. $\mathrm{conv}_{\mathit{temp}}$ denotes the 1D channel-wise temporal sub-convolution whose kernel size is 3 and $\mathrm{conv}_{\mathit{spa}}$ indicates the 3$\times$3 2D spatial sub-convolution.

In this module, the different fragments have different receptive fields. For example, the output of the first fragment $\mathbf{X}^o_1$ is the same as input fragment $\mathbf{X}_1$; thus, its receptive field is 1$\times$1$\times$1. By aggregating information from former fragments in series, the equivalent receptive field of the last fragment $\mathbf{X}^o_4$ has been enlarged three times. Finally, a simple concatenation strategy is adopted to combine multiple outputs.
\begin{equation}
	\mathbf{X}^o = \left[ \mathbf{X}^o_1;\mathbf{X}^o_2;\mathbf{X}^o_3;\mathbf{X}^o_4 \right], \quad \mathbf{X}^o \in \mathbb{R}^{N \times T \times C \times H \times W}
\end{equation}
The obtained output feature $\mathbf{X}^o$ involves spatiotemporal representations capturing different temporal ranges. It is superior to the local temporal representations obtained by using a single local convolution in typical approaches.

\begin{figure}[]
    \begin{center}
        \includegraphics[width=8cm]{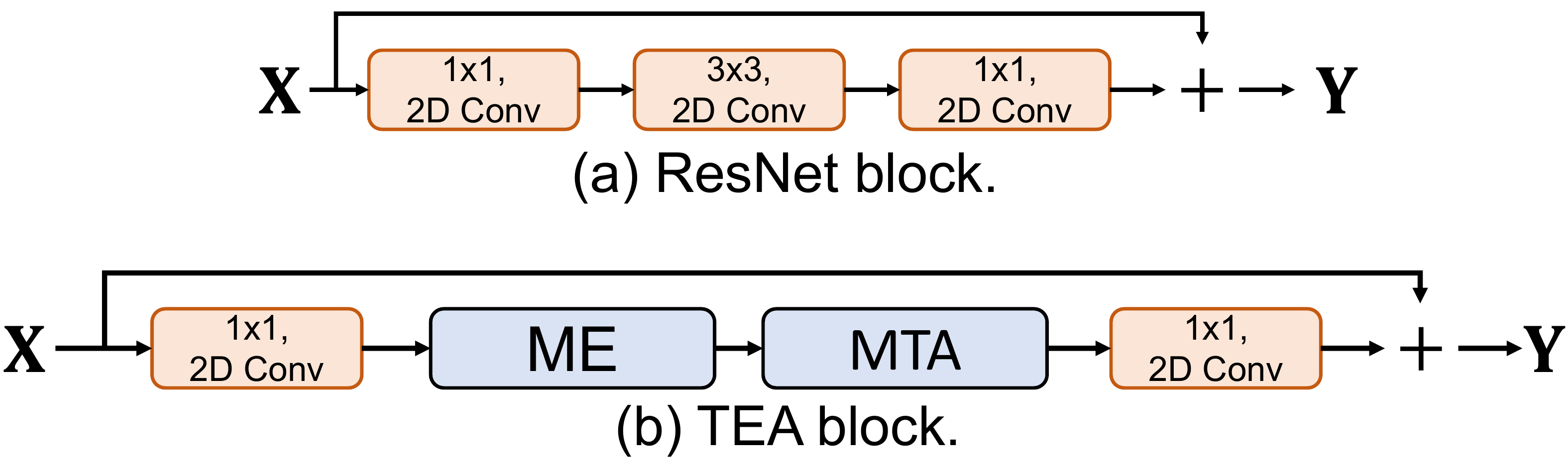}
    \end{center}
    \vspace{-1em}
        \caption{The motion excitation (ME) module is placed after the first 1$\times$1 convolution layer. The multiple temporal aggregation (MTA) module is utilized to replace the 3$\times$3 convolution layer.}
        \label{fig:resblock}
        \vspace{-1em}
    \end{figure}

\subsection{Integration with ResNet Block}
Finally, we describe how to integrate the proposed modules into standard ResNet block \cite{he2016deep} to construct our temporal excitation and aggregation (TEA) block. The approach is illustrated in Figure \ref{fig:resblock}. For computational efficiency, the motion excitation (ME) module is integrated into the residual path after the bottleneck layer (the first 1$\times$1 Conv layer). The multiple temporal aggregation (MTA) module is utilized to replace the original 3$\times$3 Conv layer in the residual path. The action recognition network can be constructed by stacking the TEA blocks.

\section{Experiments}
\subsection{Datasets}

The proposed approach is evaluated on two large-scale action recognition datasets, Something-Something V1 \cite{goyal2017something} and Kinetic400 \cite{carreira2017quo}, and other two small-scale datasets, HMDB51 \cite{kuehne2011hmdb} and UCF101 \cite{soomro2012ucf101}. As pointed in \cite{xie2018rethinking,zhou2018temporal}, most of the categories in Kinetics, HMDB, and UCF can be recognized by considering the background scene information only, and the temporal understanding is not very important in most cases. While the categories of Something-Something focus on human interactions with daily life objects, for example, ``{\itshape pull something}'' and ``{\itshape push something}''. Classifying these interactions requires more considerations of temporal information. Thus the proposed method is mainly evaluated on Something-Something since our goal is to improve the temporal modeling ability.

Kinetics contains 400 categories and provides download URL links for $\sim$240k training videos and $\sim$20k validation videos. In our experiments, we successfully collect 223,127 training videos and 18,153 validation videos, because a small fraction of the URLs (around 10\%) is no longer valid. For the Kinetics dataset, the methods are learned on the training set and evaluated on the validation set. HMDB contains 51 classes and 6,766 videos, while UCF includes 101 categories with 13,320 videos. For these two datasets, we follow TSN \cite{wang2016temporal} to utilize three different training/testing splits for evaluation, and the average results are reported.

Something-Something V1 includes 174 categories with 86,017 training videos, 11,522 validation videos, and 10,960 test videos. All of them have been split into individual frames at the same rate, and the extracted frames are also publicly available. The methods are learned on the training set and measured on the validation set and test set.

\subsection{Implementation Details}
\label{sec:exp-impl}
We utilize 2D ResNet-50 as the backbone and replace each ResNet block with the TEA block from conv2 to conv5. The sparse sampling strategy \cite{wang2016temporal} is utilized to extract $T$ frames from the video clips ($T=8$ or 16 in our experiments). During training, random scaling and corner cropping are utilized for data augmentation, and the cropped region is resized to 224$\times$224 for each frame{\footnote{More training details can be found in supplementary materials.}.

During the test, two evaluation protocols are considered to trade-off accuracy and speed. 1) {\itshape efficient protocol} (center crop$\times$1 clip), in which 1 clip with $T$ frames is sampled from the video. Each frame is resized to 256$\times$256, and a central region of size 224$\times$224 is cropped for action prediction. 2) {\itshape accuracy protocol} (full resolution$\times$10 clips), in which 10 different clips are randomly sampled from the video, and the final prediction is obtained by averaging all clips' scores. For each frame in a video clip, we follow the strategy proposed by \cite{wang2018non} and resize the shorter size to 256 with maintaining the aspect ratio. Then 3 crops of 256$\times$256 that cover the full-frame are sampled for action prediction.

\subsection{Experimental Results}
\subsubsection{Ablation Study}
In this section, we first conduct several ablation experiments to testify the effectiveness of different components in our proposed TEA block. Without loss of generality, the models are trained with 8 frames on the Something-Something V1 training set and evaluated on the validation set. Six baseline networks are considered for comparison, and their corresponding blocks are illustrated in Figure \ref{fig:baselines}. The comparison results, including the classification accuracies and inference protocols, are shown in Table \ref{tab:main-abl}.
\begin{itemize}
	\item {\bfseries (2+1)D ResNet}. In the residual branch of the standard ResNet block, a 1D channel-wise temporal convolution is inserted after the first 2D spatial convolution.
	\item {\bfseries (2+1)D Res2Net}. The channel-wise temporal convolution is integrated into Res2Net block \cite{gao2019res2net}. In Res2Net, the 3$\times$3 spatial convolution of ResNet block is deformed to be a group of sub-convolutions.
	\item {\bfseries Multiple Temporal Aggregation (MTA)}. The motion excitation module is removed from the proposed TEA network.
	\item {\bfseries Motion Excitation (ME)}. Compared with the (2+1)D ResNet baseline, the proposed motion excitation module is added to the residual path.
	\item {\bfseries (2+1)D SENet}. The SE block \cite{hu2018squeeze,hu2019squeeze} replaces the motion excitation module in the ME baseline. The SE block utilizes two fully connected layers to produce modulation weights from original features, and then apply the obtained weights to rescale the features.
	\item {\bfseries ME w/o Residual}. The residual connection is removed from the ME baseline. Thus the output feature is obtained by directly multiplying the input feature with the motion-sensitive weights, \ie, $\mathbf{X}^o = \mathbf{X} \odot \mathbf{A} $.\end{itemize}

\begin{figure}[]
    \begin{center}
        \includegraphics[width=8cm]{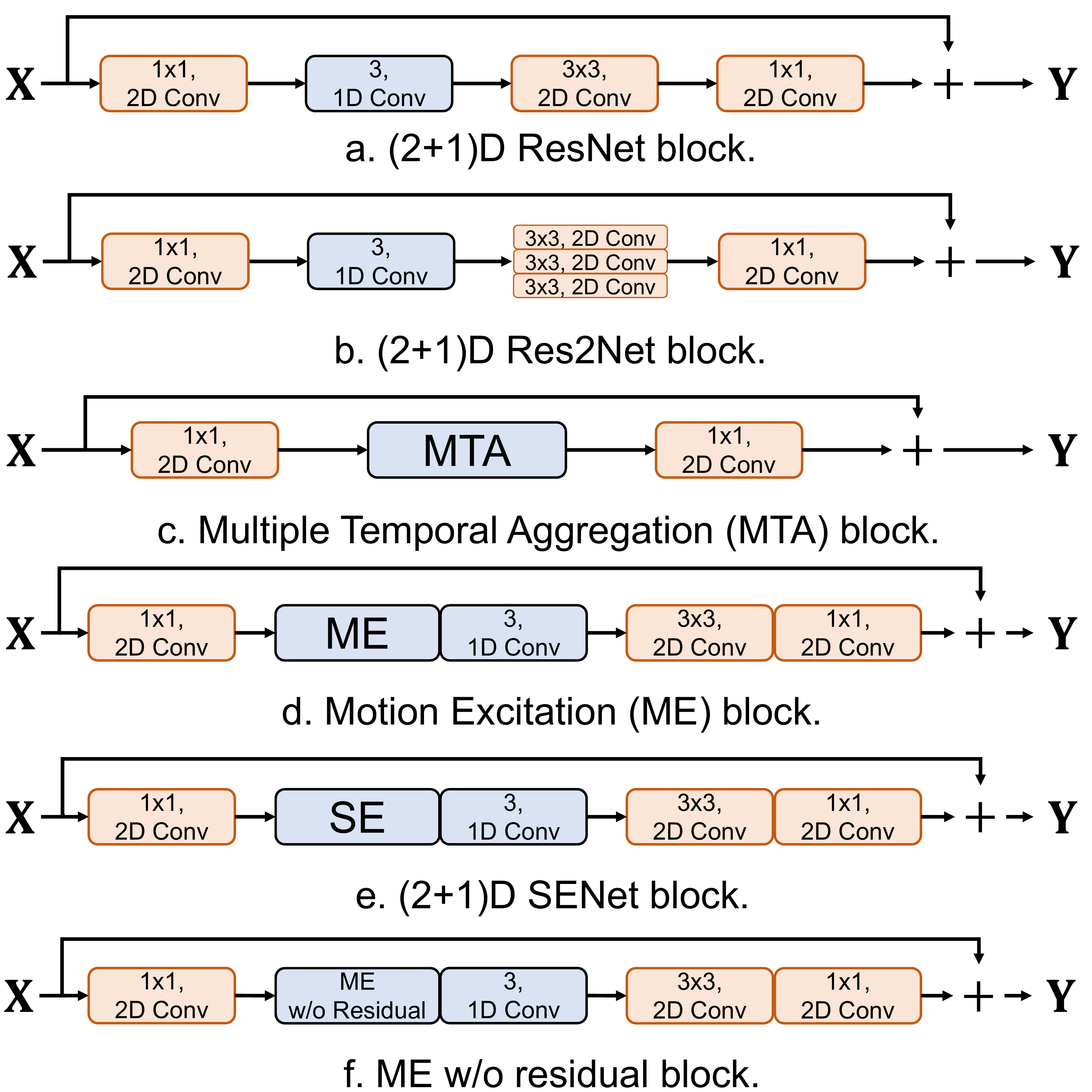}
    \end{center}
    \vspace{-1em}
        \caption{The altered blocks of different baselines based on standard ResNet block \cite{he2016deep}.}
        \label{fig:baselines}
        \vspace{-1em}
    \end{figure}

\paragraph{Effect of Multiple Temporal Aggregation.} Firstly, it can be seen from the first compartment of Table \ref{tab:main-abl} that the MTA baseline outperforms the (2+1)D ResNet baseline by a large margin (47.5\% \vs 46.0\%). Compared with the (2+1)D ResNet baseline, the capable long-range temporal aggregation can be constructed in the MTA module by utilizing the hierarchical structure to enlarge the equivalent receptive field of the temporal dimension in each block, which results in the performance improvements.

Moreover, considering the proposed MTA module enlarges both spatial and temporal receptive fields, it is thus necessary to ascertain the independent impact of the two aspects. To this end, we then compare the (2+1)D ResNet baseline with the (2+1)D Res2Net baseline. In (2+1)D Res2Net, the group of sub-convolutions is applied to spatial dimension only, and the equivalent receptive field of temporal dimension is unchanged in this model. We can see that the accuracies of the two baselines are similar and both inferior to that of MTA (46.0\%/46.2\% \vs 47.5\%). It proves that exploring complicated spatial structures and sophisticated spatial representations have, to some extent, limit impacts on the action recognition task. The key to improving the performance of action recognition is capable and reliable temporal modeling ability.

 \begin{table}[t]
      \renewcommand{\arraystretch}{1.0}
     \caption{\label{tab:main-abl} Comparison results on Something-Something. }
      \begin{center}
      \scriptsize
      \begin{tabular}{c c c c}
      \toprule
      \multicolumn{1}{c}{\bfseries Method} & \multicolumn{1}{c}{\bfseries Frames$\times$Crops$\times$Clips} & \multicolumn{1}{c}{\begin{tabular}[c]{@{}c@{}}{\bfseries Val} \\ {\bfseries Top-1 (\%)} \end{tabular}} & \multicolumn{1}{c}{\begin{tabular}[c]{@{}c@{}}{\bfseries Val} \\ {\bfseries Top-5 (\%)} \end{tabular}} \\ \midrule \midrule
      
      \multicolumn{1}{c}{(2+1)D ResNet (a)$^1$ } & \multicolumn{1}{c}{8$\times$1$\times$1} & \multicolumn{1}{c}{46.0} & \multicolumn{1}{c}{75.3} \\
      \multicolumn{1}{c}{(2+1)D Res2Net (b)$^1$ } & \multicolumn{1}{c}{8$\times$1$\times$1} & \multicolumn{1}{c}{46.2} & \multicolumn{1}{c}{75.5} \\
      \multicolumn{1}{c}{MTA (c)$^1$ } & \multicolumn{1}{c}{8$\times$1$\times$1} & \multicolumn{1}{c}{47.5} & \multicolumn{1}{c}{76.4} \\ \midrule
      \multicolumn{1}{c}{TEA } & \multicolumn{1}{c}{8$\times$1$\times$1} & \multicolumn{1}{c}{{\bfseries 48.9}} & \multicolumn{1}{c}{{\bfseries 78.1}} \\ \midrule \midrule
      \multicolumn{1}{c}{(2+1)D ResNet (a)$^1$ } & \multicolumn{1}{c}{8$\times$1$\times$1} & \multicolumn{1}{c}{46.0} & \multicolumn{1}{c}{75.3} \\
      \multicolumn{1}{c}{(2+1)D SENet (e)$^1$ } & \multicolumn{1}{c}{8$\times$1$\times$1} & \multicolumn{1}{c}{46.5} & \multicolumn{1}{c}{75.6} \\
      \multicolumn{1}{c}{ME w/o Residual (f)$^1$ } & \multicolumn{1}{c}{8$\times$1$\times$1} & \multicolumn{1}{c}{47.2} & \multicolumn{1}{c}{76.1} \\
      \multicolumn{1}{c}{STM \cite{jiang2019stm}$^2$ } & \multicolumn{1}{c}{8$\times$1$\times$1} & \multicolumn{1}{c}{47.5} & \multicolumn{1}{c}{-}  \\
       \multicolumn{1}{c}{ME (d)$^1$ } & \multicolumn{1}{c}{8$\times$1$\times$1} & \multicolumn{1}{c}{48.4} & \multicolumn{1}{c}{77.5} \\ \midrule
       \multicolumn{1}{c}{TEA } & \multicolumn{1}{c}{8$\times$1$\times$1} & \multicolumn{1}{c}{{\bfseries 48.9}} & \multicolumn{1}{c}{{\bfseries 78.1}} \\
     \bottomrule
      \end{tabular}
      \footnotesize
     \begin{tablenotes}
      \item [1] 1. {\bfseries XX} ({\bfseries y}). {\bfseries XX} indicates the XX baseline, and {\bfseries y} represents that the architecture of the corresponding block is the y-th one in Figure \ref{fig:baselines}.
      \item [2] 2. The result of STM using efficient inference protocol is cited from Table 9 in \cite{jiang2019stm}. 
      \end{tablenotes}
      \end{center}
      \vspace{-1em}
\end{table}

  \begin{table*}[!htp]
      \renewcommand{\arraystretch}{0.9}
     \caption{\label{tab:ss} Comparison results of TEA with other state-of-the-art methods on Something-Something V1. }
     \vspace{-0.5em}
      \begin{center}
      \footnotesize
      \begin{tabular}{l c c c c c c c}
      \toprule
      \multicolumn{1}{c}{\bfseries Method} & \multicolumn{1}{c}{\bfseries Backbone} & \multicolumn{1}{c}{\bfseries Frames$\times$Crops$\times$Clips} & \multicolumn{1}{c}{\bfseries FLOPs} & \multicolumn{1}{c}{\bfseries Pre-train} & \multicolumn{1}{c}{\begin{tabular}[c]{@{}c@{}}{\bfseries Val} \\ {\bfseries Top-1 (\%)}\end{tabular}} & \multicolumn{1}{c}{\begin{tabular}[c]{@{}c@{}}{\bfseries Val} \\ {\bfseries Top-5 (\%)}\end{tabular}} & \multicolumn{1}{c}{\begin{tabular}[c]{@{}c@{}}{\bfseries Test} \\ {\bfseries Top-1 (\%)}\end{tabular}} \\ \midrule
      \midrule
      \multicolumn{1}{l}{3D/(2D+3D) CNNs:} \\
      \multicolumn{1}{c}{I3D-RGB \cite{wang2018videos} } & \multicolumn{1}{c}{3D ResNet50} & \multirow{3}{*}{32$\times$3$\times$2} & \multicolumn{1}{c}{153G$\times$3$\times$2} & \multirow{3}{*}{\begin{tabular}[c]{@{}c@{}}ImgNet\\ +\\ K400\end{tabular}} & \multicolumn{1}{c}{41.6} & \multicolumn{1}{c}{72.2} & \multicolumn{1}{c}{-} \\
      \multicolumn{1}{c}{NL I3D-RGB \cite{wang2018videos} } & \multicolumn{1}{c}{3D ResNet50} &  & \multicolumn{1}{c}{168G$\times$3$\times$2} &  & \multicolumn{1}{c}{44.4} & \multicolumn{1}{c}{76.0} & \multicolumn{1}{c}{-} \\
      \multicolumn{1}{c}{NL I3D+GCN-RGB \cite{wang2018videos} } & \multicolumn{1}{c}{3D ResNet50+GCN} &  & \multicolumn{1}{c}{303G$\times$3$\times$2} &  & \multicolumn{1}{c}{46.1} & \multicolumn{1}{c}{76.8} & \multicolumn{1}{c}{45.0} \\
      \midrule
      \multicolumn{1}{c}{ECO-RGB \cite{zolfaghari2018eco} } & \multirow{3}{*}{BNIncep+3D Res18} & \multicolumn{1}{c}{8$\times$1$\times$1} & \multicolumn{1}{c}{32G$\times$1$\times$1} & \multirow{3}{*}{K400} & \multicolumn{1}{c}{39.6} & \multicolumn{1}{c}{-} & \multicolumn{1}{c}{-} \\
      \multicolumn{1}{c}{ECO$_{En}$-RGB \cite{zolfaghari2018eco} } &  & \multicolumn{1}{c}{92$\times$1$\times$1} & \multicolumn{1}{c}{267G$\times$1$\times$1} &  & \multicolumn{1}{c}{46.4} & \multicolumn{1}{c}{-} & \multicolumn{1}{c}{42.3} \\
      
      \multicolumn{1}{c}{\textcolor{gray}{ECO$_{En}$-(RGB+Flow) \cite{zolfaghari2018eco}} } &  & \multicolumn{1}{c}{\textcolor{gray}{92 + 92}} & \multicolumn{1}{c}{\textcolor{gray}{N/A$^2$}} &  & \multicolumn{1}{c}{\textcolor{gray}{49.5}} & \multicolumn{1}{c}{\textcolor{gray}{-}} &  \multicolumn{1}{c}{\textcolor{gray}{43.9}} \\
      \midrule 
      \midrule
      \multicolumn{1}{l}{2D/(2+1)D CNNs:} \\
      \multicolumn{1}{c}{TSN-RGB \cite{wang2016temporal} } & \multicolumn{1}{c}{BNInception} & \multirow{2}{*}{8$\times$1$\times$1} & \multicolumn{1}{c}{16G$\times$1$\times$1} & \multirow{2}{*}{ImgNet} & \multicolumn{1}{c}{19.5} & \multicolumn{1}{c}{-} & \multicolumn{1}{c}{-} \\
      \multicolumn{1}{c}{TSN-RGB \cite{wang2016temporal} } & \multicolumn{1}{c}{ResNet50} &  & \multicolumn{1}{c}{33G$\times$1$\times$1} &  & \multicolumn{1}{c}{19.7} & \multicolumn{1}{c}{-} & \multicolumn{1}{c}{-} \\
      \midrule
      \multicolumn{1}{c}{STM-RGB \cite{jiang2019stm} } & \multirow{2}{*}{ResNet50} & \multicolumn{1}{c}{8$\times$3$\times$10} & \multicolumn{1}{c}{33G$\times$3$\times$10} & \multirow{2}{*}{ImgNet} & \multicolumn{1}{c}{49.2} & \multicolumn{1}{c}{79.3} & \multicolumn{1}{c}{-} \\
      \multicolumn{1}{c}{STM-RGB \cite{jiang2019stm} } &  & \multicolumn{1}{c}{16$\times$3$\times$10} & \multicolumn{1}{c}{67G$\times$3$\times$10} & & \multicolumn{1}{c}{50.7} & \multicolumn{1}{c}{80.4} & \multicolumn{1}{c}{43.1} \\ \midrule 
      \multicolumn{1}{c}{TSM-RGB \cite{lin2019tsm} } & \multirow{4}{*}{ResNet50} & \multicolumn{1}{c}{8$\times$1$\times$1} & \multicolumn{1}{c}{33G$\times$1$\times$1} & \multirow{4}{*}{\begin{tabular}[c]{@{}c@{}}ImgNet\\ +\\ K400\end{tabular}} & \multicolumn{1}{c}{43.4} & \multicolumn{1}{c}{73.2} & \multicolumn{1}{c}{-} \\
      \multicolumn{1}{c}{TSM-RGB \cite{lin2019tsm} } &  & \multicolumn{1}{c}{16$\times$1$\times$1} & \multicolumn{1}{c}{65G$\times$1$\times$1} &  & \multicolumn{1}{c}{44.8} & \multicolumn{1}{c}{74.5} & \multicolumn{1}{c}{-} \\
      \multicolumn{1}{c}{TSM$_{en}$-RGB \cite{lin2019tsm} } & & \multicolumn{1}{c}{8 + 16} & \multicolumn{1}{c}{33G + 65G} & & \multicolumn{1}{c}{46.8} & \multicolumn{1}{c}{76.1} & \multicolumn{1}{c}{-} \\
      \multicolumn{1}{c}{\textcolor{gray}{TSM-(RGB+Flow) \cite{lin2019tsm}} } & & \multicolumn{1}{c}{\textcolor{gray}{16 + 16}} & \multicolumn{1}{c}{\textcolor{gray}{N/A$^2$}} & & \multicolumn{1}{c}{\textcolor{gray}{50.2}} & \multicolumn{1}{c}{\textcolor{gray}{79.5}} & \multicolumn{1}{c}{{\bfseries \textcolor{gray}{47.0}}} \\
      \midrule
      \midrule
      \multicolumn{1}{c}{TEA (Ours) } & \multirow{4}{*}{ResNet50} & \multicolumn{1}{c}{8$\times$1$\times$1} & \multicolumn{1}{c}{35G$\times$1$\times$1} & \multirow{4}{*}{ImgNet} & \multicolumn{1}{c}{48.9} & \multicolumn{1}{c}{78.1} & \multicolumn{1}{c}{-} \\
      \multicolumn{1}{c}{TEA (Ours) } &  & \multicolumn{1}{c}{8$\times$3$\times$10} & \multicolumn{1}{c}{35G$\times$3$\times$10} & & \multicolumn{1}{c}{51.7} & \multicolumn{1}{c}{80.5} & \multicolumn{1}{c}{45.3} \\
      \multicolumn{1}{c}{TEA (Ours) } &  & \multicolumn{1}{c}{16$\times$1$\times$1} & \multicolumn{1}{c}{70G$\times$1$\times$1} &  & \multicolumn{1}{c}{51.9} & \multicolumn{1}{c}{80.3} & \multicolumn{1}{c}{-} \\
      \multicolumn{1}{c}{TEA (Ours) } &  & \multicolumn{1}{c}{16$\times$3$\times$10} & \multicolumn{1}{c}{70G$\times$3$\times$10} & & \multicolumn{1}{c}{\textbf{52.3}} & \multicolumn{1}{c}{\textbf{81.9}} & \multicolumn{1}{c}{46.6} \\ \bottomrule
      \end{tabular}
      \footnotesize
     \begin{tablenotes}
      \item 1. ``ImgNet'' denotes ImageNet dataset \cite{deng2009imagenet,russakovsky2015imagenet} and ``K400'' indicates Kinetics400 datasets \cite{carreira2017quo}.
      \item 2.  ``N/A'' represents that the FLOPs cannot be accurately measured because of extracting optical flow.
      \end{tablenotes}
      \end{center}
      \vspace{-1.5em}
\end{table*}

 \paragraph{Effect of Motion Modeling.} To testify the effectiveness of the motion modeling for action recognition, we compare the ME baseline with the (2+1)D ResNet baseline. In the second compartment of Table \ref{tab:main-abl}, we can see that the action recognition performance is significantly increased by considering the motion encoding (48.1\% \vs 46.0\%). The discovery of motion-sensitive features will force the networks to focus on dynamic information that reflects the actual actions. 
  
 To prove that such improvement is not brought by introducing extra parameters and soft attention mechanisms, we then compare the (2+1)D SENet baseline with the (2+1)D ResNet baseline. (2+1)D SENet adds the SE block at the start of the trunk path, aiming to excite the informative feature channels. However, the SE block is applied to each frame of videos independently, and the temporal information is not considered in this approach. Thus, the performance of the (2+1)D SENet baseline is similar to the (2+1)D ResNet baseline (46.5\% \vs 46.0\%). The improvement is quite limited.
 
 Finally, we explore several designs for motion modeling. We first compare the ME baseline with the ME w/o Residual baseline. It can be seen that the performance decreases from 48.1\% to 47.2\% without residual connections since the static information related background scenes will be eliminated in ME w/o Residual. It proves that the scene information is also beneficial for action recognition, and the residual connection is necessary for the motion excitation module. Then we compare the ME baseline with STM \cite{jiang2019stm}. We can see that ME attains higher accuracy than STM (48.4\% \vs 47.5\%), which verifies the excitation mechanism utilized in the proposed method is superior to the simple {\itshape add} approach used in STM. When additionally considering the long-range temporal relationship by introducing the MTA module, the accuracy of our method (TEA) can be further improved to 48.9\%.

\subsubsection{Comparisons with the State-of-the-arts}

In this section, we first compare TEA with the existing state-of-the-art action recognition methods on Something-Something V1 and Kinetics400. The comprehensive statistics, including the classification results, inference protocols, and the corresponding FLOPs, are shown in Table \ref{tab:ss} and \ref{tab:kinetics}.

In both tables, the first compartment contains the methods based on 3D CNNs or the mixup of 2D and 3D CNNs, and the methods in the second compartment are all based on 2D or (2+1)D CNNs. Due to the high computation costs of 3D CNNs, the FLOPs of methods in the first compartment are typically higher than others. Among all existing methods, the most efficient ones are TSN$_{8f}$ \cite{wang2016temporal} and TSM$_{8f}$ \cite{lin2019tsm} with only 33G FLOPs. Compared with these methods, the FLOPs of our proposed TEA network slightly increases to 35G (1.06$\times$), but the performance is increased by a big margin, a relative improvement of 5.4 \% (48.8\% \vs 43.4\%).

\begin{table*}[htp]
      \renewcommand{\arraystretch}{0.9}
      \caption{\label{tab:kinetics} Comparison results of TEA with other state-of-the-art methods on Kinetics400 validation set. }
      \vspace{0.5em}
      \begin{center}
      \footnotesize
      \begin{tabular}{l c c c c c c}
      \toprule
      \multicolumn{1}{c}{\bfseries Method} & \multicolumn{1}{c}{\bfseries Backbone} & \multicolumn{1}{c}{\bfseries Frames$\times$Crops$\times$Clips} & \multicolumn{1}{c}{\bfseries FLOPs} & \multicolumn{1}{c}{\bfseries Pre-train} & \multicolumn{1}{c}{\bfseries Top-1 (\%)} & \multicolumn{1}{c}{\bfseries Top-5 (\%)} \\ \midrule \midrule
      \multicolumn{1}{l}{3D/(2D+3D) CNNs:} \\
      \multicolumn{1}{c}{I3D-RGB \cite{carreira2017quo} } & \multirow{2}{*}{Inception V1} & \multirow{2}{*}{64$\times$N/A$\times$N/A$^1$} & \multicolumn{1}{c}{108G$\times$N/A$\times$N/A} & \multicolumn{1}{c}{ImgNet} & \multicolumn{1}{c}{72.1} & \multicolumn{1}{c}{90.3} \\
      \multicolumn{1}{c}{I3D-RGB \cite{carreira2017quo} } &  &  & \multicolumn{1}{c}{108G$\times$N/A$\times$N/A} & \multicolumn{1}{c}{None} & \multicolumn{1}{c}{67.5} & \multicolumn{1}{c}{87.2} \\ \midrule
      
       \multicolumn{1}{c}{ECO-RGB$_{En}$ \cite{zolfaghari2018eco} } & \multicolumn{1}{c}{BNIncep+3D Res18} & \multicolumn{1}{c}{92$\times$1$\times$1} & \multicolumn{1}{c}{267G$\times$1$\times$1} & \multicolumn{1}{c}{None} & \multicolumn{1}{c}{70.0} & \multicolumn{1}{c}{-} \\ \midrule
      
      
      \multicolumn{1}{c}{NL I3D-RGB \cite{wang2018videos} } & \multicolumn{1}{c}{3D ResNet101} & \multicolumn{1}{c}{32$\times$6$\times$10} & \multicolumn{1}{c}{359G$\times$6$\times$10} & ImgNet & \multicolumn{1}{c}{77.7} & \multicolumn{1}{c}{93.3} \\ \midrule

      \multicolumn{1}{c}{NL SlowFast \cite{feichtenhofer2019slowfast}}
      & \multicolumn{1}{c}{3D ResNet101} & \multicolumn{1}{c}{(16+8)$\times$3$\times$10} & \multicolumn{1}{c}{234G$\times$3$\times$10} & \multicolumn{1}{c}{None} & \multicolumn{1}{c}{\textbf{79.8}} & \multicolumn{1}{c}{\textbf{93.9}} \\
  \midrule \midrule
      \multicolumn{1}{l}{2D/(2+1)D CNNs:} \\
      \multicolumn{1}{c}{TSN-RGB \cite{wang2016temporal} } & \multicolumn{1}{c}{BNInception} & \multirow{2}{*}{25$\times$10$\times$1} & \multicolumn{1}{c}{53G$\times$10$\times$1} & \multirow{2}{*}{ImgNet} & \multicolumn{1}{c}{69.1} & \multicolumn{1}{c}{88.7} \\
      \multicolumn{1}{c}{TSN-RGB \cite{wang2016temporal} } & \multicolumn{1}{c}{Inception v3} &  & \multicolumn{1}{c}{80G$\times$10$\times$1} &  & \multicolumn{1}{c}{72.5} & \multicolumn{1}{c}{90.2} \\ \midrule
      
      \multicolumn{1}{c}{R(2+1)D \cite{tran2018closer} } & \multicolumn{1}{c}{ResNet-34} & \multicolumn{1}{c}{32$\times$1$\times$10} & \multicolumn{1}{c}{152G$\times$1$\times$10} & \multicolumn{1}{c}{None} & \multicolumn{1}{c}{72.0} & \multicolumn{1}{c}{90.0} \\ \midrule
      
      \multicolumn{1}{c}{STM-RGB \cite{jiang2019stm} } & \multicolumn{1}{c}{ResNet50} & \multicolumn{1}{c}{16$\times$3$\times$10} & \multicolumn{1}{c}{67G$\times$3$\times$10} & \multicolumn{1}{c}{ImgNet} & \multicolumn{1}{c}{73.7} & \multicolumn{1}{c}{91.6} \\ \midrule
      
      \multicolumn{1}{c}{TSM-RGB \cite{lin2019tsm} } & \multirow{2}{*}{ResNet50} & \multicolumn{1}{c}{8$\times$3$\times$10} & \multicolumn{1}{c}{33G$\times$3$\times$10} & \multirow{2}{*}{ImgNet} & \multicolumn{1}{c}{74.1} & \multicolumn{1}{c}{-} \\
      \multicolumn{1}{c}{TSM-RGB \cite{lin2019tsm} } &  & \multicolumn{1}{c}{16$\times$3$\times$10} & \multicolumn{1}{c}{65G$\times$3$\times$10} &  & \multicolumn{1}{c}{74.7} & \multicolumn{1}{c}{-} 
      \\ \midrule \midrule
      
      \multicolumn{1}{c}{TEA (Ours) } & \multirow{4}{*}{ResNet50} & \multicolumn{1}{c}{8$\times$1$\times$1} & \multicolumn{1}{c}{35G$\times$1$\times$1} & \multirow{4}{*}{ImgNet} & \multicolumn{1}{c}{72.5} & \multicolumn{1}{c}{90.4} \\
      \multicolumn{1}{c}{TEA (Ours) } &  & \multicolumn{1}{c}{8$\times$3$\times$10} & \multicolumn{1}{c}{35G$\times$3$\times$10} & & \multicolumn{1}{c}{75.0} & \multicolumn{1}{c}{91.8} \\
      \multicolumn{1}{c}{TEA (Ours) } &  & \multicolumn{1}{c}{16$\times$1$\times$1} & \multicolumn{1}{c}{70G$\times$1$\times$1} &  & \multicolumn{1}{c}{74.0} & \multicolumn{1}{c}{91.3} \\
      \multicolumn{1}{c}{TEA (Ours) } &  & \multicolumn{1}{c}{16$\times$3$\times$10} & \multicolumn{1}{c}{70G$\times$3$\times$10} & & \multicolumn{1}{c}{76.1} & \multicolumn{1}{c}{92.5} \\ \bottomrule
      \end{tabular}
      \footnotesize
      \begin{tablenotes}
      \item 1. ``ImgNet'' denotes ImageNet dataset \cite{deng2009imagenet,russakovsky2015imagenet} and ``None'' indicates training models from scratch.
      \item 2. ``N/A'' represents that the authors do not report the inference protocol in their paper.
      \end{tablenotes}
      \end{center}
      \vspace{-1em}
\end{table*}

The superiority of our TEA on Something-Something is quite impressive. It confirms the remarkable ability of TEA for temporal modeling. Using efficient inference protocol (center crop$\times$1 clip) and 8 input frames, the proposed TEA obtains 48.8\%, which significantly outperforms TSN and TSM with similar FLOPs (19.7\%/43.4\%). This results even exceeds the ensemble result of TSM, which combines the two models using 8 and 16 frames, respectively (TSM$_{En}$, 46.8\%). When utilizing 16 frames as input and applying a more laborious accuracy evaluation protocol (full resolution$\times$10 clips), the FLOPs of our method increase to $\sim$2000G, which is similar to NL I3D+GCN \cite{wang2018videos}. But the proposed method significantly surpasses NL I3D+GCN and all other existing methods (52.3\% \vs 46.1\%) on the validation set. Our performance on the test set (46.6\%) also outperforms most of the existing methods. Moreover, we do not require additional COCO images \cite{lin2014microsoft} to pre-train an object detector as in \cite{wang2018videos}. When compared with the methods utilizing both RGB and optical flow modalities, \ie, ECO$_{En}$-(RGB+Flow) \cite{zolfaghari2018eco} (49.5\%) and TSM-(RGB+Flow) \cite{lin2019tsm} (50.2\%), the obtained result (52.3\%) also shows substantial improvements.

On Kinetics400, the performance of our method (76.1\%) is inferior to that of SlowFast \cite{feichtenhofer2019slowfast} (79.8\%). However, the SlowFast networks adopt the deeper networks (ResNet101) based on 3D CNNs and utilize time-consuming non-local \cite{wang2018non} operations. When comparing methods with similar efficiency, such as TSM \cite{lin2019tsm} and STM \cite{jiang2019stm}, TEA obtains better performance. When adopting 8 frames as input, TEA gains $\sim$1\% higher accuracy than TSM (75.0\% \vs 74.1\%). While utilizing 16 input frames, our TEA method outperforms both TSM$_{16f}$ and STM $_{16f}$ with a large margin (76.1\% \vs 74.7\%/73.7\%).

Finally, we report comparison results on HMDB51 and UCF101 in Table \ref{tab:small}. Our method achieves 73.3\% on HMDB51 and 96.9\% on UCF101 with the accuracy inference protocol. The performance of our model (TEA$_{16f}$) outperforms most of the existing methods except for I3D \cite{carreira2017quo}. I3D is based on 3D CNNs and additional input modality; thus, its computational FLOPs will be far more than ours.

 \begin{table}[t ]
      \renewcommand{\arraystretch}{1.0}
     \caption{\label{tab:small} Comparison results on HMDB51 and UCF101.}
      \begin{center}
      \footnotesize
      \begin{tabular}{c c c c}
      \toprule
      \multicolumn{1}{c}{\bfseries Method}  & \multicolumn{1}{c}{ {\bfseries Backbone} } & \multicolumn{1}{c}{\begin{tabular}[c]{@{}c@{}}{\bfseries HMDB51} \\ {\bfseries MCA (\%)$^1$}\end{tabular}} & \multicolumn{1}{c}{\begin{tabular}[c]{@{}c@{}} {\bfseries UCF101} \\ {\bfseries MCA (\%)$^1$}\end{tabular}} \\ \midrule \midrule
      \multicolumn{1}{c}{\textcolor{gray}{I3D-(RGB+Flow) \cite{carreira2017quo}}} & \multicolumn{1}{c}{\textcolor{gray}{3D Inception}} & 
      \multicolumn{1}{c}{{\bfseries \textcolor{gray}{80.7}}} & \multicolumn{1}{c}{ {\bfseries \textcolor{gray}{98.0}}} \\
      \multicolumn{1}{c}{\textcolor{gray}{TSN-(RGB+Flow) \cite{wang2016temporal}}} & \multicolumn{1}{c}{\textcolor{gray}{BNInception}} & \multicolumn{1}{c}{\textcolor{gray}{68.5}} & \multicolumn{1}{c}{\textcolor{gray}{94.0}} \\
      \multicolumn{1}{c}{StNet \cite{he2019stnet}} & \multicolumn{1}{c}{ResNet50}  & \multicolumn{1}{c}{-} & \multicolumn{1}{c}{93.5} \\
       \multicolumn{1}{c}{TSM$^2$} & \multicolumn{1}{c}{ResNet50}  & \multicolumn{1}{c}{70.7} & \multicolumn{1}{c}{94.5} \\
      \multicolumn{1}{c}{STM \cite{jiang2019stm}} & \multicolumn{1}{c}{ResNet50} & \multicolumn{1}{c}{72.2} & \multicolumn{1}{c}{96.2} \\ \midrule \midrule
       \multicolumn{1}{c}{TEA (Ours)} & \multicolumn{1}{c}{ResNet50} & \multicolumn{1}{c}{73.3} & \multicolumn{1}{c}{96.9} \\
     \bottomrule
      \end{tabular}
      \footnotesize
      \begin{tablenotes}
      \item 1. MCA denotes mean class accuracy.
      \item 2. TSM does not report MCA results, and the listed results are cited from STM \cite{jiang2019stm}.
      \end{tablenotes}	

      \end{center}
      \vspace{-1em}
\end{table}
 
\section{Conclusion}
In this paper, we propose the Temporal Excitation and Aggregation (TEA) block, including the motion excitation (ME) module and the multiple temporal aggregation (MTA) module for both short- and long-range temporal modeling. Specifically, the ME module could insert the motion encoding into the spatiotemporal feature learning approach and enhance the motion pattern in spatiotemporal features. In the MTA module, the reliable long-range temporal relationship can be established by deforming the local convolutions into a group of sub-convolutions to enlarge the equivalent temporal receptive field. The two proposed modules are integrated into the standard ResNet block and cooperate for capable temporal modeling.

\section{Ackonwledgement}
This work is supported by the Video Understanding Middle Platform of the Platform and Content Group (PCG) at Tencent. The authors would like to thank Wei Shen for his helpful suggestions.

{\small
\bibliographystyle{ieee_fullname}
\bibliography{egbib}
}
\clearpage
\begin{appendices}
\section{Temporal Convolutions in TEA}
For video action recognition tasks, previous works typically adopt 3D convolutions to simultaneously model spatial and temporal features or utilize (2+1)D convolutions to decouple the temporal representation learning and the spatial feature modeling. The 3D convolutions will bring tremendous computations costs and preclude the benefits of utilizing ImageNet pre-training. Moreover, the blend of spatial and temporal modeling also makes the model harder to optimize. Thus, as shown in Figure 2 of the main text, in our proposed TEA module, we choose (2+1)D architectures and adopt separated 1D temporal convolutions to process temporal information. To train (2+1)D models, a straightforward approach is fine-tuning the 2D spatial convolutions from ImageNet pre-trained networks meanwhile initialize the parameters of 1D temporal convolutions with random noise.

However, according to our observations, it will lead to contradictions to simultaneously optimize the temporal convolutions and spatial convolutions in a unified framework because the temporal information exchange between frames brought by temporal convolutions might harm the spatial modeling ability. In this section, we will describe a useful tactic to deal with this problem for effectively optimizing temporal convolutions in video recognition models.

Before introducing the tactic, we first retrospect the recently proposed action recognition method TSM \cite{lin2019tsm}. Different from previous works adopting temporal convolutions \cite{tran2018closer,xie2018rethinking}, TSM utilizes an ingenious {\bfseries shift} operator to endow the model with the temporal modeling ability without introducing any parameters into 2D CNN backbones. Concretely, given an input feature $\mathbf{X} \in \mathbb{R}^{N \times T \times C \times H \times W}$, the shift operations are denoted to shift the feature channels along the temporal dimension. Suppose the input feature is a five-dimensional tensor, the example pseudo-codes of left/right operator are as follows:
\begin{equation}
\begin{array}{cc}
	\mathbf{X}\left[ n, t, c, h, w \right] = \mathbf{X}\left[ n, t+1, c, h, w \right], & \quad 1 \leq t \leq T-1 \\
	\mathbf{X}\left[ n, t, c, h, w \right] = \mathbf{X}\left[ n, t-1, c, h, w \right], & \quad 2 \leq t \leq T 
\end{array}
\end{equation}

By utilizing such {\itshape shift} operation, the spatial features at time step $t$, $\mathbf{X}\left[ :, t, :, :, :\right]$, achieves temporal information exchange between neighboring time steps, $t-1$ and $t+1$. In practice, the shift operation can be conducted on all or some of the feature channels, and the authors explore several shift options \cite{lin2019tsm}. Finally, they find that if all or most of the feature channels are shifted, {\bfseries the performance will decrease due to worse spatial modeling ability}.

The possible reason for this observation is that TSM utilizes 2D CNN backbone pre-trained on ImageNet as initializations to fine-tune the models on new video datasets. The benefit of utilizing such a pre-training approach is that the features of pre-trained ImageNet models would contain some kind of useful spatial representations. But after shifting a part of feature channels to neighboring frames, such useful spatial representations modeled by the shifted channels are no longer accessible for current frame. The newly obtained representations become the combination of three successive frames, \ie, $t-1$, $t$ and $t+1$, which are disordered and might be ``meaningless'' for the current frame $t$.

To balance these two aspects, TSM experimentally chooses to shift {\bfseries a small part} of feature channels. More specifically, the first 1/8 channels are shifted left, the second 1/8 channels are shifted right, and the last 3/4 channels are fixed. Formally,

\begin{widetext}
\begin{equation}
\label{equ:shift}
\begin{array}{lccr}
	\mathbf{X}\left[ n, t, c, h, w \right] = \mathbf{X}\left[ n, t+1, c, h, w \right], & \quad 1 \leq t \leq T-1, & \quad 1 \leq c \leq C/8 & \textit{left shift} \\
	\mathbf{X}\left[ n, t, c, h, w \right] = \mathbf{X}\left[ n, t-1, c, h, w \right], & \quad 2 \leq t \leq T, & \quad C/8 < c \leq C/4 & \textit{right shift} \\
	\mathbf{X}\left[ n, t, c, h, w \right] = \mathbf{X}\left[ n, t, c, h, w \right], & \quad 1 \leq t \leq T, & \quad C/4 < c \leq C & \textit{unchanged}
\end{array}
\end{equation}
\end{widetext}
This {\itshape part shift} operation has been proved effective in TSM and obtains impressive action recognition accuracies on several benchmarks.

When we think over the shift operation proposed by TSM, we find that it is actually a {\bfseries special case} of general 1D temporal convolutions and we will show an example to illustrate this. Instead of conducting shift operation on input feature $\mathbf{X}$ as in Equation \ref{equ:shift}, we aim to utilize a 1D channel-wise temporal convolution to achieve the same control for $\mathbf{X}$ (we utilize channel-wise convolution for simplicity, and the formulas in Equation \ref{equ:conv} can be extended to general convolutions.). Concretely, a 1D channel-wise temporal convolution is conducted on input features, whose kernel size is 3, the kernel weights at the shifted channels are set to fixed $\left[ 0,0,1\right]$ or $\left[ 1,0,0 \right]$ and the kernel weights at unchanged channels are set to fixed $\left[ 0,1,0 \right]$. Formally,
\begin{widetext}
\begin{equation}
\label{equ:conv}
\begin{array}{lccr}
    \mathbf{X}_{\mathit{reshape}} = \mathrm{Reshape}(\mathbf{X}), & \mathbf{X} \in \mathbb{R}^{N \times T \times C \times H \times W}, & \mathbf{X}_{\mathit{reshape}} \in \mathbb{R}^{NHW \times C \times T} & \textit{reshape \& permute} \\
	\mathbf{X}_{\mathit{shift}} = \mathbf{K}*\mathbf{X}_{\mathit{reshape}}, & \mathbf{X}_{\mathit{shift}} \in \mathbb{R}^{NHW \times C \times T}, & \mathbf{K} \in \mathbb{R}^{C \times 1 \times 3} & \textit{temporal convolution} \\
	\mathbf{K}\left[ {\mathit cout}, 1, k \right] = 1, & 1 \leq cout \leq C/8, & k=3 & \textit{left shift} \\
	\mathbf{K}\left[ {\mathit cout}, 1, k \right] = 0, & 1 \leq cout \leq C/8, & k=1,2 & \textit{left shift} \\
	\mathbf{K}\left[ {\mathit cout}, 1, k \right] = 1, & C/8 < cout \leq C/4, & k=1 & \textit{right shift} \\
	\mathbf{K}\left[ {\mathit cout}, 1, k \right] = 0, & C/8 < cout \leq C/4, & k=2,3 & \textit{right shift} \\
	\mathbf{K}\left[ {\mathit cout}, 1, k \right] = 1, & C/4 < cout \leq C, & k=2 & \textit{unchanged} \\
	\mathbf{K}\left[ {\mathit cout}, 1, k \right] = 0, & C/4 < cout \leq C, & k=1,3 & \textit{unchanged} \\
	
\end{array}
\end{equation}
\end{widetext}
where $\mathbf{K}$ denotes convolutional kernels, and $*$ indicates the convolution operation. It's not hard to see that the Equation \ref{equ:conv} is totally equivalent to Equation \ref{equ:shift}. The shift operation proposed in TSM can be considered as a 1D temporal convolution operation with {\itshape fixed pre-designed} kernel weights. Thus, one natural question is, {\itshape whether the performance of video action recognition can be improved by relaxing the fixed kernel weights to learnable kernel weights}. We experiment to verify this question.
\begin{table}[t]
      \renewcommand{\arraystretch}{1.0}
     \small
     \caption{\label{tab:conv} Comparison results on Something-Something. All the models are trained with 8 input frames, and the one clip-one crop protocol is utilized for inference.}
      \begin{center}
      \begin{tabular}{c c c}
      \toprule
      \multicolumn{1}{c}{\bfseries Method}  & \multicolumn{1}{c}{\bfseries Top-1 (\%)} & \multicolumn{1}{c}{\bfseries Top-5 (\%)} \\ \midrule \midrule
      \multicolumn{1}{c}{TSM \cite{lin2019tsm} } & \multicolumn{1}{c}{43.4} & \multicolumn{1}{c}{73.2} \\ \midrule
      \multicolumn{1}{c}{(2+1)D ResNet-Conv (Ours) } & \multicolumn{1}{c}{23.5} & \multicolumn{1}{c}{45.8} \\
      \multicolumn{1}{c}{(2+1)D ResNet-CW (Ours) } & \multicolumn{1}{c}{43.6} & \multicolumn{1}{c}{73.4} \\
      \multicolumn{1}{c}{(2+1)D ResNet-Shift (Ours) } & \multicolumn{1}{c}{\textbf{46.0}} & \multicolumn{1}{c}{\textbf{75.3}} \\
     \bottomrule
      \end{tabular}
      \end{center}
\end{table}

The experiment is conducted based on the (2+1)D ResNet baseline. The detailed descriptions of the baseline are introduced in Section 4.3.1 of the main text. We design several variants of (2+1)D ResNet, and the only difference between these variants is the type of utilized 1D temporal convolution.
\begin{itemize}
	\item {\bfseries (2+1)D ResNet-Conv} which adopts general 1D temporal convolutions. The parameters of temporal convolutions are randomly initialized.
	\item {\bfseries (2+1)D ResNet-CW} which utilizes channel-wise temporal convolutions. The parameters are also randomly initialized.
	\item {\bfseries (2+1)D ResNet-Shift}. In this variant, the channel-wise temporal convolutions are also utilized, but the parameters of the temporal convolutions are initialized as in Equation \ref{equ:conv} to perform like {\itshape part shift} operators at the beginning of the model learning.
\end{itemize}

During training, the parameters of temporal convolutions in all the three variants are learnable, and the final obtained models are evaluated on Something-Something V1 with 8 frames as input and the efficient inference protocol.

The comparison results are shown in Table \ref{tab:conv}. We first notice that when comparing the (2+1)D ResNet-Conv with (2+1)D ResNet-CW, the (2+1)D ResNet-Conv baseline fails to obtain acceptable performance. As we mentioned in the main text, different channels of spatial features capture different information; thus, the temporal combination of each channel should be different and learned independently. Moreover, the general temporal convolution will introduce lots of parameters and make the model harder to be optimized.

The second observation is that the performance of (2+1)D ResNet-CW is only slightly higher than that of TSM (43.6\% \vs 43.4\%). Although the learnable kernel weights endow the model with the ability to learn dynamic temporal information exchange patterns, all the features channels are disarrayed with randomly initialized convolutions. It finally results in damage to the spatial feature learning capacity and counters the benefits of effective temporal representation learning.

Inspired by the {\itshape part shift} strategy utilized in TSM, (2+1)D ResNet-Shift proposes to initialize the temporal convolutions to perform as {\itshape part shift}, which grantees the spatial feature learning ability inheriting from the pre-trained ImageNet 2D CNN models. Meanwhile, along with the optimization of the models, the temporal convolutions can gradually explore more effective temporal information aggregation strategy with learnable kernel weights. Finally, this {\itshape part shift} initialization strategy obtains 46.0\% top-1 accuracy, which is substantially higher than TSM.

According to the experiment, we can see that by drawing lessons from TSM and designing a {\itshape part shift} initialization strategy, the performance of action recognition can be improved by using 1D temporal convolutions with learnable kernel weights. This strategy is thus applied to each of the temporal convolutions in the proposed TEA module.

\section{Training Details}
In this section, we will elaborate on detailed configurations for training the TEA network on different datasets. The codes and related experimental logs will be made publicly available soon.

\begin{table*}[htp]
\renewcommand{\arraystretch}{1.0}
\caption{\label{tab:infer} Comparison results on Something-Something V1.}
\begin{center}	
\begin{tabular}{l c c c}
\toprule
Method & Frame$\times$Crops$\times$Clips & Inference Time (ms/v)  & Val Top-1 (\%) \\ \midrule \midrule
TSN (2D ResNet) [41]   & 8$\times$1$\times$1 & 0.0163              & 19.7   \\
TSM [25]               & 8$\times$1$\times$1 & 0.0185            & 43.4   \\
TSM [25]               & 16$\times$1$\times$1 & 0.0359              & 44.8   \\
STM [20]               & 8$\times$1$\times$1 & 0.0231              & 47.5   \\
I3D (3D ResNet) [43]   & 32$\times$3$\times$2 & 4.4642                & 41.6  \\ \midrule

ME (d in Figure 4)     & 8$\times$1$\times$1 &  0.0227            & 48.4   \\
MTA (c in Figure 4)    & 8$\times$1$\times$1 & 0.0256             & 47.5   \\
TEA                    & 8$\times$1$\times$1 &  0.0289            & \textbf{48.9}   \\ \bottomrule
\end{tabular}
\end{center}
\end{table*}

\subsection{Model Initializations}
Following the previous action recognition works \cite{wang2016temporal,lin2019tsm,jiang2019stm}, we utilize 2D CNNs pre-trained on ImageNet dataset as the initializations for our network. Notice that the proposed multiple temporal aggregation (MTA) module is based on Res2Net \cite{gao2019res2net}, whose architecture is different from the standard ResNet \cite{he2016deep}. We thus select the released Res2Net50 model ($\mathrm{res2net50\_26w\_4s}$\footnote{\url{https://shanghuagao.oss-cn-beijing.aliyuncs.com/res2net/res2net50_26w_4s-06e79181.pth}}) pre-trained on ImageNet to initialize the proposed network.

Although Res2Net has been proved a stronger backbone than ResNet on various image-based tasks in \cite{gao2019res2net}, \eg, image classification, and image object detection, it will {\bfseries NOT} brings many improvements for video action recognition task. As we have discussed in the ablation study section (Section 4.3.1) of the main text, the temporal modeling ability is the key factor for video-based tasks, rather than the complicated spatial representations. The experimental results in Table 1 of the main text also verify this. With more powerful 2D backbones, the action recognition performance of (2+1)D Res2Net only obtains slight improvements over (2+1)D ResNet (46.2\% \vs 46.0\%).

\subsection{Hyperparameters}
Most of the experimental settings are as the same as TSM \cite{lin2019tsm} and STM \cite{jiang2019stm}. For experiments on Kinetics and Something-Something V1 \& V2, the networks are fine-tuned from ImageNet pre-trained models. All the batch normalization layers \cite{ioffe2015batch} are enabled during training. The learning rate and weight decay of the classification layer (a fully connected layer) are set to 5$\times$ higher than other layers. For Kinetics, the batch size, initial learning rate, weight decay, and dropout rate are set to 64, 0.01, 1e-4, and 0.5 respectively; for Something-Something, these hyperparameters are set to 64, 0.02, 5e-4 and 0.5 respectively. For these two datasets, the networks are trained for 50 epochs using stochastic gradient descent (SGD), and the learning rate is decreased by a factor of 10 at 30, 40, and 45 epochs.

When fine-tuning Kinetics models on other small datasets, \ie, HMDB51 \cite{kuehne2011hmdb} and UCF101 \cite{soomro2012ucf101}, the batch normalization layers are frozen except the first one following TSN \cite{wang2016temporal}. The batch size, initial learning, weight decay and dropout rate are set to 64, 0.001, 5e-4 and 0.8 for both the two datasets. The learning rate and weight decay of the classification layer are set to 5$\times$ higher than other layers. The learning rate is decreased by a factor of 10 at 10 and 20 epochs. The training procedure stops at 25 epochs.

Finally, the learning rate should match the batch size as suggested by \cite{goyal2017accurate}. For example, the corresponding learning rate should increase two times if the batch size scales up from 64 to 128.

\section{The Effect of the Transformation Convolutions in the ME Module}
When calculating feature-level motion representations in the ME module, we first apply a channel-wise transformation convolution on features at the time step $t+1$. The reason is that motions will cause spatial displacements for the same objects between two frames, and it will result in mismatched motion representation to directly compute differences between displaced features. To address this issue, we add a 3$\times$3 convolution at time step $t+1$ attempting to capture the matched regions of the same object from contexts. According to our verification, this operation leads to further improvement of TEA on Something-Something V1 (from 48.4\% to 48.9\%). Moreover, we found that conducting transformation on both $t$ and $t+1$ time steps does not improve the performance but introduces more operations.

 \begin{table*}[htp]
      \renewcommand{\arraystretch}{1.0}
     \caption{\label{tab:location} Comparison results on Something Something V1.}
      \begin{center}
      \begin{tabular}{c c c c c c c}
      \toprule
      \multicolumn{1}{c}{\bfseries Stage} & \multicolumn{1}{c}{\bfseries Backbone} & \multicolumn{1}{c}{\begin{tabular}[c]{@{}c@{}}{\bfseries Number of } \\ {\bfseries the TEA Blocks}\end{tabular}} & \multicolumn{1}{c}{ {\bfseries Frames$\times$Crops$\times$Clips} } & \multicolumn{1}{c}{\begin{tabular}[c]{@{}c@{}}{\bfseries Val} \\ {\bfseries Top1 (\%)}\end{tabular}} & \multicolumn{1}{c}{\begin{tabular}[c]{@{}c@{}}{\bfseries Val} \\ {\bfseries Top5 (\%)}\end{tabular}} \\ \midrule \midrule
      \multicolumn{1}{c}{conv2} & \multicolumn{1}{c}{ResNet50} & \multicolumn{1}{c}{3} & \multicolumn{1}{c}{8$\times$1$\times$1} & \multicolumn{1}{c}{43.5} & \multicolumn{1}{c}{72.2} \\
      \multicolumn{1}{c}{conv3} & \multicolumn{1}{c}{ResNet50} & \multicolumn{1}{c}{4} & \multicolumn{1}{c}{8$\times$1$\times$1} &\multicolumn{1}{c}{45.3} & \multicolumn{1}{c}{74.5} \\
      \multicolumn{1}{c}{conv4} & \multicolumn{1}{c}{ResNet50} & \multicolumn{1}{c}{6} & \multicolumn{1}{c}{8$\times$1$\times$1} &\multicolumn{1}{c}{47.1} & \multicolumn{1}{c}{76.2} \\
      \multicolumn{1}{c}{conv5} & \multicolumn{1}{c}{ResNet50} & \multicolumn{1}{c}{3} & \multicolumn{1}{c}{8$\times$1$\times$1} &\multicolumn{1}{c}{46.7} & \multicolumn{1}{c}{75.8} \\ \midrule \midrule
       \multicolumn{1}{c}{TEA (conv2$\sim$conv5)} & \multicolumn{1}{c}{ResNet50} & \multicolumn{1}{c}{16} & \multicolumn{1}{c}{8$\times$1$\times$1} & \multicolumn{1}{c}{{\bfseries 48.9}} & \multicolumn{1}{c}{{\bfseries 78.1}} \\
     \bottomrule
      \end{tabular}
      \footnotesize
      \end{center}
\end{table*}

\section{Runtime Analysis}
We show the accuracies and inference times of TEA and other methods in Table \ref{tab:infer}. All these tests are conducted on one P40 GPU, and the batch size is set to 16. The time for data loading is excluded from the evaluation. Compared with STM, TEA achieves higher accuracy with similar efficiency. Compared with TSM$_{16F}$ and I3D, both the effectivity and efficiency of TEA$_{8F}$ are superior. The runtime of the \textbf{2D ResNet baseline} (TSN) is nearly 1.8x faster than TEA. But its performance is \textbf{far behind} ours (19.7\% \vs 48.9\%).

We further analyze the efficiency of each component in TEA by comparing TEA with MTA and ME, respectively. We can see that the hierarchical stages in MTA cause an increase of $0.0062s$ ($\triangle t$$=$TEA$-$ME), as the multiple stages need to be sequentially processed. The increased time brought by ME is $0.0033s$ ($\triangle t$$=$TEA$-$MTA). Please note that for an input feature $\mathbf{X}$ with $T$ timestamps, it is not required to subtract between adjacent features \textit{timestamp-wise} and then concatenate $T$-1 differences. We only need to \textit{slice} $\mathbf{X}$ along the temporal dimension \textit{twice} to obtain the features of time 1$\sim$$T$-1 and time 2$\sim$$T$ respectively. Then only one subtraction is performed to obtain the final feature differences. The example pseudo codes are as follows, and the time cost of this approach is only $0.0002$$s$.

\begin{lstlisting}
# x (input features): NxTxCxHxW
f_t, __ = x.split([T-1,1], dim=1)
__, f_t1 = x.split([1,T-1], dim=1)
# diff_f (feature differences): Nx(T-1)xCxHxW
diff_f = f_t1 - f_t
\end{lstlisting}


\section{The Location of the TEA Block}
As described in Section 4.2 of the main text, the TEA blocks are utilized to replace all the ResNet blocks of the ResNet-50 backbone from conv2 to conv5. In this section, we conduct an ablation study to explore the different impacts caused by inserting the TEA blocks into ResNet at different locations. Specifically, we replace all the ResNet blocks with the TEA blocks at a particular stage, \eg, conv2, and leave all other stages, \eg, conv3$\sim$conv5, unchanged. The networks are learned on the training set of Something-Something V1 and measured on its validation set. During the test, the efficient protocol (center crop$\times$1 clip) is adopted, and the comparison results are shown in Table \ref{tab:location}. It can be seen that, in general, the action recognition performance of inserting TEA blocks into the later stages (\ie, conv4/conv5, 47.1\%/46.7\%) is superior to that of inserting the TEA blocks into the early stages (\ie, conv2/conv3, 43.5\%/45.3\%). The spatiotemporal features at the later stage would capture temporal information from a larger range and realize capable temporal aggregations. Thus, the TEA blocks at the later stages would have more effective and determinative impacts for improving temporal modeling ability, which finally results in higher action recognition performance. When inserting the TEA blocks into all stages of the ResNet backbone, the performance of our method further increases and achieves the best result (48.9\%).

\begin{figure}[t]
    \begin{center}
        \includegraphics[width=9.0cm]{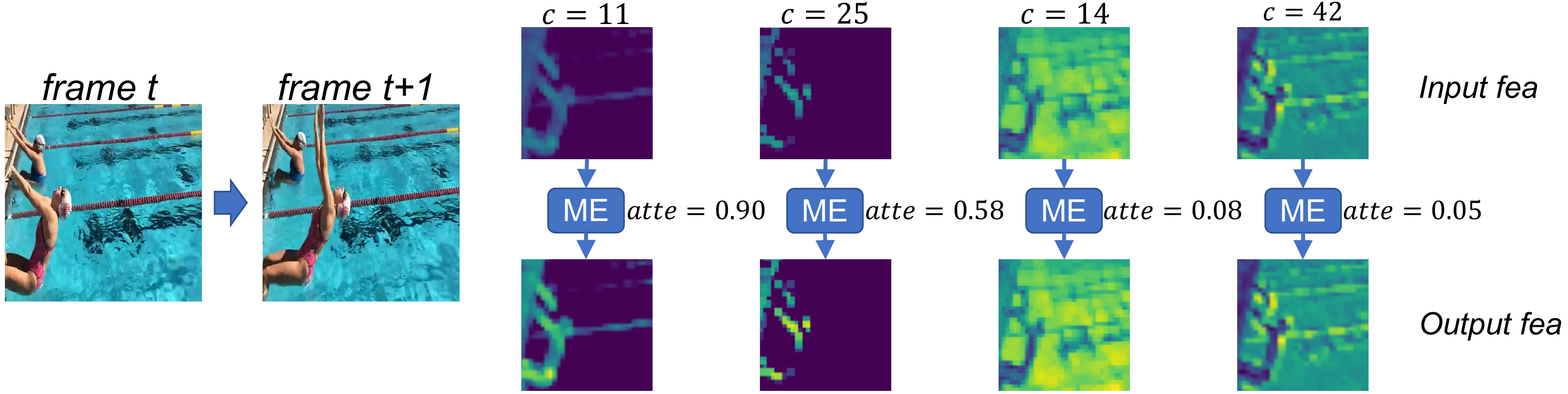}
    \end{center}
        \caption{Visualizations of the input and output features of ME in Conv2\_1 block.}
        \label{fig:vis}
    \end{figure}

\section{The verification for the assumption of ME}

To verify the assumption of ME, we give a visualization example in Figure \ref{fig:vis}. We can see that different feature channels capture different information. For example, on channels 11 and 25, features model the moving swimmers, and the ME module enhances this motion information by giving a large attention weight ($\mathbf{A}$$=$0.90/0.58). In contrast, on channels 14 and 42, the background information is simply preserved with a quite lower attention weight, 0.08/0.05.


\begin{table*}[t]
      \renewcommand{\arraystretch}{1.0}
     \caption{\label{tab:sthv2} Comparison results on Something Something V2.}
      \begin{center}
      \begin{tabular}{c c c c c c c}
      \toprule
      \multicolumn{1}{c}{\bfseries Method} & \multicolumn{1}{c}{\bfseries Backbone} & \multicolumn{1}{c}{ {\bfseries Frames$\times$Crops$\times$Clips} } & \multicolumn{1}{c}{\begin{tabular}[c]{@{}c@{}}{\bfseries Val} \\ {\bfseries Top1 (\%)}\end{tabular}} & \multicolumn{1}{c}{\begin{tabular}[c]{@{}c@{}}{\bfseries Val} \\ {\bfseries Top5 (\%)}\end{tabular}} & \multicolumn{1}{c}{\begin{tabular}[c]{@{}c@{}}{\bfseries Test} \\ {\bfseries Top1 (\%)}\end{tabular}} & \multicolumn{1}{c}{\begin{tabular}[c]{@{}c@{}} {\bfseries Test} \\ {\bfseries Top5 (\%)}\end{tabular}} \\ \midrule \midrule
      \multicolumn{1}{c}{TSN-RGB$^1$} & \multicolumn{1}{c}{ResNet50} & \multicolumn{1}{c}{16$\times$3$\times$2} & \multicolumn{1}{c}{30.0} & \multicolumn{1}{c}{60.5} & \multicolumn{1}{c}{-} & \multicolumn{1}{c}{-} \\
      \multicolumn{1}{c}{TSM-RGB \cite{lin2019tsm}} & \multicolumn{1}{c}{ResNet50} & \multicolumn{1}{c}{16$\times$3$\times$2} & \multicolumn{1}{c}{59.4} & \multicolumn{1}{c}{86.1} & \multicolumn{1}{c}{60.4} & \multicolumn{1}{c}{87.3} \\
      \multicolumn{1}{c}{STM \cite{jiang2019stm}} & \multicolumn{1}{c}{ResNet50} & \multicolumn{1}{c}{16$\times$3$\times$10} & \multicolumn{1}{c}{64.2} & \multicolumn{1}{c}{89.8} & \multicolumn{1}{c}{63.5} & \multicolumn{1}{c}{89.6} \\ \midrule \midrule
       \multicolumn{1}{c}{TEA} & \multicolumn{1}{c}{ResNet50} & \multicolumn{1}{c}{16$\times$3$\times$10} & \multicolumn{1}{c}{{\bfseries 65.1}} & \multicolumn{1}{c}{{\bfseries 89.9}} & \multicolumn{1}{c}{63.2} & \multicolumn{1}{c}{{\bfseries 89.7}} \\
     \bottomrule
      \end{tabular}
      \footnotesize
     \begin{tablenotes}
      \item 1. The results of TSN \cite{wang2016temporal} on Something-Something V2 are cited from the implementation of TSM \cite{lin2019tsm}.
      \end{tablenotes}
      \end{center}
\end{table*}


\section{Experimental Results on Something-Something V2}
In this section, we compare the proposed TEA network with other state-of-the-art methods on Something-Something V2 \cite{goyal2017something}. Something-Something V2 is a newer release version of the Something-Something dataset. It contains 168,913 training videos, 24,777 validation videos and 27,157 videos. Its size is twice larger than Something-Something V1 (108,499 videos in total). The TEA network is learned on the training set and evaluated on the validation set and test set. The accuracy inference protocol (full resolution$\times$10 clips) is utilized for evaluation, and the results are shown in Table \ref{tab:sthv2}. We can see that on the validation set, our result (65.1\%) outperforms those of the existing state-of-the-art methods. On the test set, the obtained number is also comparable to the state-of-the-art result (63.2\% \vs 63.5\%). These results verify the effectiveness of the proposed TEA network on Something-Something V2.

\end{appendices}

\end{document}